\documentclass{isprs} 
\usepackage{setspace}
\usepackage{amssymb}
\usepackage{graphicx}
\usepackage{geometry} 
\usepackage{epstopdf}
\usepackage{amsmath}
\usepackage{booktabs}
\usepackage{multirow}
\usepackage{subcaption}
\usepackage{mathtools}
\usepackage{xcolor}
\usepackage{float}
\usepackage{microtype}
\usepackage{hyperref}
\hypersetup{
    colorlinks=true,      
    allcolors=black,      
    urlcolor=blue         
}
\usepackage[labelsep=period]{caption}  
\usepackage[british]{babel} 
\usepackage[hang]{footmisc}

\newcommand{\PAR}[1]{\noindent {\bf #1~}}
\begin{document}

\title{Shape Representation using Gaussian Process mixture models}
\date{}


\author{
Panagiotis Sapoutzoglou, George Terzakis, Georgios Floros, Maria Pateraki}

\address{
	 Laboratory of Photogrammetry, School of Rural, Surveying and Geoinformatics Engineering, \\National Technical University of Athens, Greece \\ (psapoutzoglou, gterzakis, gfloros, mpateraki)@mail.ntua.gr\\
}




\abstract{
Traditional explicit 3D representations, such as point clouds and meshes, demand significant storage to capture fine geometric details and require complex indexing systems for surface lookups, making functional representations an efficient, compact, and continuous alternative. In this work, we propose a novel, object-specific functional shape representation that models surface geometry with Gaussian Process (GP) mixture models. Rather than relying on computationally heavy neural architectures, our method is lightweight, leveraging GPs to learn continuous directional distance fields from sparsely sampled point clouds. We capture complex topologies by anchoring local GP priors at strategic reference points, which can be flexibly extracted using any structural decomposition method (e.g. skeletonization, distance-based clustering). Extensive evaluations on the ShapeNetCore and IndustryShapes datasets demonstrate that our method can efficiently and accurately represent complex geometries.}

\maketitle

\section{Introduction}

 Shape representation constitutes a fundamental problem in a variety of fields such as 3D computer vision, photogrammetry, and graphics. Compact and accurate representations of 3D shapes are essential for tasks ranging from object modeling, reconstruction, and 6D pose estimation to the creation of digital twins, large-scale urban modeling, and heritage documentation. In these photogrammetric workflows, the transition from raw sensor data to structured surfaces is often a critical bottleneck. Standard outputs, which typically consist of massive, unstructured point clouds, suffer from noise, occlusion, and varying point densities, creating a burgeoning need for representations that bridge the gap between discrete data and continuous geometry. While explicit geometric representations (e.g., point clouds, meshes, and voxels) can capture fine levels of detail, they lack storage parsimony, leading to high memory costs and requiring complex indexing systems for quick surface lookups. On the other hand, functional models (implicit representations) are typically parametric, highly storage-efficient, and adept at learning complex topologies.
Consequently, a line of research represents surfaces implicitly, either as a continuous volumetric
field or as the decision boundary of a classifier.
Other approaches use neural rendering to learn the object representation, optimizing a neural network, a sparse
grid of spherical harmonics or an unstructured set of
3D Gaussians. Local kernel functions emerged as a building block for implicit shape representations. Overall, neural shape representations bear significant benefits, such as generalization and storage efficiency.
These models are optimized to capture the statistical characteristics of an entire object class, facilitating tasks like shape completion, rather than reconstructing the high fidelity, idiosyncratic details of a single, specific geometry.

In this work we propose an object-specific implicit representation: Functional modeling of surface geometry using Gaussian Processes (GPs). 
GPs are prominent tools for non-parametric regression~\cite{Rasmussen2004}. 
In contrast to neural models, our method leverages the ability of GPs to model continuous functions from irregularly sparse sampled data and apply this concept in the context of a probabilistic model that learns the shape of an object as the mixture of multiple directional distance fields anchored at reference points specially placed in the object's skeletal outline. Unlike methods that utilize directional distance fields (DDFs) along with a neural architecture~\cite{Tretschk2020,Armstrong2022,feng2022,Yenamandra2024}, we abstain from using any neural component. Our GP approach draws from the work by Has~\cite{has2023gradient} on classifier mixtures and exploits it for shape representation. 
The resulting mixture model ensures geometric continuity and sparsity, capturing finer shape detail while avoiding the heavy training burden associated with deep implicit methods.
The contributions of our method are two-fold.
\begin{enumerate}
    \item We present a novel, functional shape representation as a mixture of GP based priors. This representation is ``lightweight'' as it completely abstains from deep neural components while operating solely on sparsely sampled surface points.
    \item We evaluate the representation capabilities of the proposed model using both synthetic and real data and compare it against state-of-the-art competitors.
\end{enumerate}

The source code is available at \url{https://github.com/POSE-Lab/GP-mixture-shape-representation}.

\section{Related Work}

 \subsection{Shape representation}
Existing approaches for representing 3D geometry can be broadly divided into distinct paradigms based on how they mathematically model an object's surface~\cite{cremers2015image}. Point-clouds, meshes, and voxels are the most popular \textit{explicit} representations able to capture arbitrary amounts of detail in the objects' shapes. They are used either as a standalone model or in conjunction with an encoder-decoder architecture~\cite{achlioptas2017learning,Wang2018_pixel2mesh,Brock2016}. Typical shortcomings of these representations include increased storage requirements for higher detail levels, and the need for special lookup structures to support operations such as nearest neighbor queries. 

On the other hand, \textit{implicit} representations are typically parametric functional models that are more storage-efficient and have proven to be adept at learning complex topologies~\cite{Farshian2023}. A line of research attempts to implicitly represent the object's surface, either as a continuous volumetric field~\cite{park2019deepsdf} or as the decision boundary of a classifier~\cite{Mescheder2019}. Other approaches use neural rendering to learn the object representation, optimizing a neural network~\cite{Mindenhall_NeRF}, a sparse grid of spherical harmonics~\cite{fridovich2022plenoxels} or an unstructured set of 3D Gaussians~\cite{kerbl20233d}. Local kernel functions~\cite{williams2021nkf,huang2023nksr} have also been used as a building block for implicit shape representations, focusing on the problem of surface reconstruction from sparse point clouds. Several extensions have been proposed, including the use of hierarchical structures~\cite{Paschalidou2020LearningUH} and the deformation of the underlying model~\cite{zheng2021deep}.
Overall, neural shape representations bear significant benefits, such as generalization, continuity in the input domain and storage efficiency. However, they are not purposed to reproduce details of specific shapes but rather as more general prediction frameworks. As such, they require swaths of data for training, rendering them unappealing for generating templates. 
On the contrary, switching to primitive representations  has its own drawbacks such as increased storage to level of detail in the shape or computationally intensive lookup queries~\cite{Farshian2023,Liao2018}.
Although Gaussian processes are prominent tools for non-parametric regression~\cite{titsias2009,williams2006gaussian}, to the best of our knowledge, they have not been utilized as a primary functional representation for 3D shapes. The closest related work by Dragiev et al.~\cite{dragiev2011} employs them specifically to refine shape uncertainty by integrating sensor data incrementally. More broadly, GPs have been extensively used for pattern learning from sparse data. Recently, neural GP ensembles~\cite{damianou2013deep} and GP-inspired architectures~\cite{garnelo2018,wilson2016} have been proposed, boasting superior performance on tasks such as orientation estimation from face images or digit magnitude prediction, rather than full surface modeling.

 \subsection{Directional distance representation}
 Directional distance representation methods have a long history in computer vision~\cite{rosenfeld1968}. Aumentado-Armstrong et al. (2022)~\nocite{Armstrong2022} exploits DDFs as a 3D representation framework. For a  given shape it learns a field that maps any position and orientation to visibility and distance, allowing differentiable rendering within an implicit architecture. Rather than modeling a volumetric field (as in NeRFs) or an explicit mesh, DDFs use a continuous field of distances around objects, facilitating their use in neural networks for 3D scene reconstruction and other inverse graphics tasks. The paper outlines some notable limitations of the DDF approach, which relate to a) rendering complexity as they demand multiple network evaluations per pixel, making real-time applications challenging, b) requirement of high-quality depth and visibility data, and c) handling occlusions as DDFs require fine control over surface visibility, potentially leading to inaccuracies. Feng et al. (2022)~\nocite{feng2022} leverages an implicit shape global representation, which models the object from the ray-based perspective, rather than 3D-point-based, demonstrating improved efficiency and quality over standard approaches like SDF~\cite{park2019deepsdf}. The shape is represented by a single network without any spatial partitions as in PatchNets~\cite{Tretschk2020}. Though due to its dependency on known views for training it may require additional modifications to produce reliable outputs, while the evaluation cost increases in high-resolution settings and certain self-intersecting geometries remain challenging.
 The method of \cite{Yenamandra2024} also utilizes a DDF representation and combines it with the SDF to accelerate 3D shape reconstruction tasks, enabling rendering with only a single evaluation per ray compared to traditional SDF-based methods that require multiple evaluations per ray for rendering. The method is optimized for speed with a simpler DDF model which limits its ability to handle occlusion and discontinuities, while \cite{Armstrong2022} focuses on handling discontinuities more accurately but with additional computational overhead. In comparison to the above methods that focus on the task of differentiable rendering in inverse graphics tasks our GP-based approach aims to handle complex shapes with multiple surface intersections by distributing reference points within the object towards a method-independent pose confidence framework for real-time applications. Thus each method’s approach reflects its intended application.
\section{Method}

\begin{figure*}[t]
    \centering
    \includegraphics[width=\textwidth]{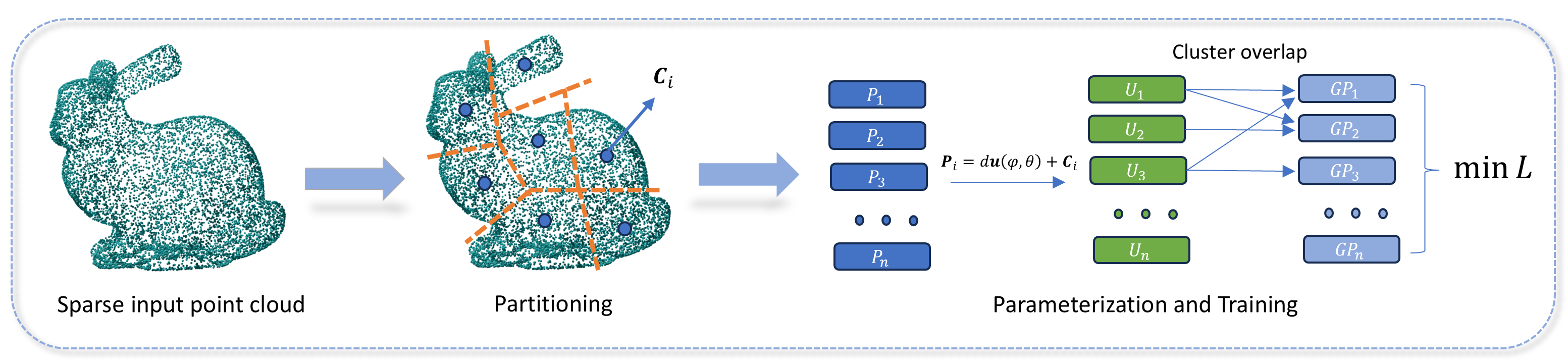}
\caption{We partition a sparse point cloud using distance-based clustering and extract reference points (Sec.~\ref{ssect:lightweighht_shape_templates}). Each 3D point $P_i$ is parameterized as a bearing vector in spherical coordinates $U_i$ relative to its reference point $C_i$. The directions, with distances as targets, serve as inputs to a Gaussian Process ($GP_i$) (Sec.~\ref{ssect:GP_prior_for_distance}). The GP mixture model forms our shape representation.}
\label{fig:Method_cobra}
 
\end{figure*}

We construct our shape representation by learning continuous directional distance fields (DDFs) from sparsely sampled point clouds. Unlike traditional methods that rely on neural networks, our approach leverages Gaussian Processes (GPs) to model these fields as a functional representation. Our method is summarized in Fig. \ref{fig:Method_cobra}. The shape is initially partitioned into regions and subsequently, special-purpose reference points are computed via skeletonization or distance-based clustering. 
The 3D points belonging to each region are then encoded as a directional distance fields along in spherical coordinates with respect to a chosen reference point. The encoding of distance fields is achieved by training local GP models that comprise our GP mixture model.


\subsection{Parametrization of Directional Distance Fields}
\label{ssect:lightweighht_shape_templates}
\begin{figure}[t]
     \centering
     \includegraphics[scale=0.15]{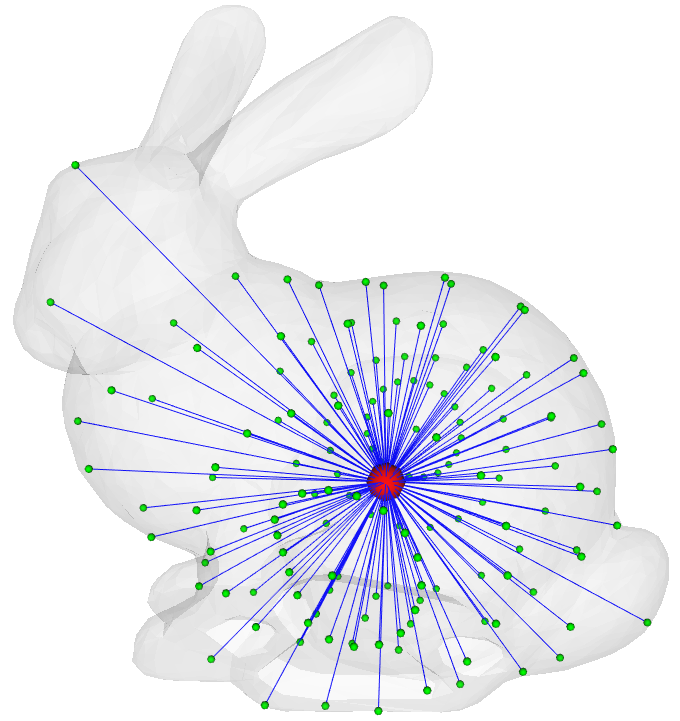}
     \caption{Spherical directional distance field (blue rays) centered at reference point $\pmb{C}$ (red).} 
     \label{fig:bunny_rays}
\end{figure}

We construct our shape representations by learning DDFs on spherical coordinate domains affixed to suitable reference points in the object's local coordinate frame. Consider a reference point $\pmb{C}$ in the object's local frame and a given point $\pmb{P}$ on the object's surface. We parameterize $\pmb{r}=\pmb{P}-\pmb{C}$ in terms of its length and bearing (see Fig.~\ref{fig:bunny_rays}) as follows:
\begin{equation}
    \pmb{r}=\pmb{P}-\pmb{C} \coloneq \left(\phi, \theta,d \right),
    \label{eq:parametrization}
\end{equation}
where $\phi, \theta$ are the spherical coordinates of the bearing vector  $\pmb{r}/\Vert\pmb{r}\Vert$ and $d=\Vert \pmb{r}\Vert$ is the Euclidean norm of $\pmb{r}$. Thus, if $\pmb{u}(\phi, \theta)  \in \mathbb{R}^3$ is the bearing vector defined by the spherical coordinates $\phi$ and $\theta$, we may obtain $\pmb{P}$ as,
\begin{equation}
    \pmb{P}=\pmb{r}+\pmb{C}=d\cdot\pmb{u}(\phi, \theta) + \pmb{C}.
    \label{eq:point_from_parameters}
\end{equation}
Thus, for a given reference point, $\pmb{C}$, in the interior of an object represented by a closed surface, we may represent all or part of the surface as a collection of distances along every possible direction from $\pmb{C}$.

\subsection{Gaussian Processes for distance prediction}
\label{ssect:GP_prior_for_distance}
Consider an object with an exterior surface that comprises points reachable by means of ray-casting from a point, $\pmb{C}$, in the interior of the object. To fit this surface to a function, we model distances from $\pmb{C}$ to the surface along a given direction $\pmb{u}(\phi, \theta)$, as a Gaussian process~\cite{Rasmussen2004,bishop2006pattern,perez2013gaussian},
\begin{equation}
    d \sim \mathcal{GP}\left(0, k\right), \;\;\text{s.t.}\;\; k : \left([0, \pi]\times[0, 2\pi]\right)^2\rightarrow \mathbb{R},
    \label{eq:gaussian_process}
\end{equation}
where $k$ is the so-called covariance function mapping any two direction parameter vectors $\pmb{\psi}=(\phi, \theta)$ and $\pmb{\psi}^{\prime}=(\phi^{\prime}, \theta^{\prime})$ to a real number, such that the Gram matrix $\pmb{K}=\left[k(\pmb{\psi}_i, \pmb{\psi}_j)\right]$ is positive semi-definite (PSD) for any finite collection $ \left\{\pmb{\psi}_1, \dots, \pmb{\psi}_n \right\}$ of direction parameter vectors~\cite{Rasmussen2004}. Different choices for the kernel $k$ give rise to different GPs. 
The Rational Quadratic (RQ) kernel, known for its capacity to regulate the extent of multi-scale behavior, stands out as a more adaptive option (see Sec.~\ref{sec:res_abl}) when compared to the Radial Basis Function (RBF) kernel \cite{shawe2004kernel,williams2006gaussian} : 


\begin{equation}
k_\mathrm{RQ}(\pmb{\psi}_i, \pmb{\psi}_j) = \left(1+\frac{d_{ij}^2}{2\alpha l^2}\right)^{-\alpha}. 
\label{eq:rational_quadratic}
\end{equation}
In this context $l > 0$, $\alpha > 0$ are hyperparameters and $d_{ij}$ denotes the Euclidean distance, $\Vert\pmb{\psi}_i-\pmb{\psi}_j\Vert$, between the orientation parameters. Perhaps a more suitable choice for $d_{ij}$ would be the geodesic distance between the bearing vectors $\pmb{u}(\pmb{\psi}_i)$, $\pmb{u}(\pmb{\psi}_j)$, but that would lead to a non-PSD kernel function as shown in \cite{kernels}, to which a  close valid alternative  would be the Euclidean distance, $\Vert \pmb{u}(\pmb{\psi}_i)- \pmb{u}(\pmb{\psi}_j)\Vert$. 

For a sufficiently sized training set of $n$ 3D points on the surface of the object, we may train the GP model to explicitly predict surface shape in terms of distance for given query directions from the point $\pmb{C}$. 
Thus, via the GP prior, for a query direction parameter vector $\pmb{x}\in\mathbb{R}^2$, we can obtain a conditional distribution over the distance to the surface along the given direction, and a collection of surface (training) data, $(\pmb{\Psi}, \pmb{d})=\left( \left[\pmb{\psi}_1,\dots, \pmb{\psi}_M\right], \left[d_1, \dots, d_M\right]\right)$:
\begin{equation}
    q\left(d\vert\ \pmb{x}, \pmb{C};(\pmb{\Psi}, \pmb{d})\right) \sim \mathcal{N}\left(\mu_{d\vert\pmb{x}}, \sigma^2_{d\vert\pmb{x}}\right),
    \label{eq:distance_likelihood}
\end{equation}
where, 
\begin{equation}
\begin{split}
\mu_{d\vert\pmb{x}}&=\pmb{K}(\pmb{x}, \pmb{\Psi})\pmb{K}(\pmb{\Psi}, \pmb{\Psi})^{-1}\pmb{d}, \\ 
    \sigma^2_{d\vert\pmb{x}}&=k(\pmb{x}, \pmb{x})-\pmb{K}(\pmb{x}, \pmb{\Psi})\pmb{K}(\pmb{\Psi}, \pmb{\Psi})^{-1}\pmb{K}(\pmb{\Psi}, \pmb{x}).
\end{split}
\label{eq:likelihood_mean_and_covariance}
\end{equation}
To discharge notation, we will henceforth omit the training data from the likelihood expression, $q\left(d\vert\ \pmb{x}, \pmb{C}; (\pmb{\Psi}, d)\right)$, and simply use,  $q\left(d\vert\ \pmb{x}, \pmb{C}\right)$.

\begin{figure}[b]
  \centering

  \begin{subfigure}{0.20\columnwidth}
    \includegraphics[width=\linewidth]{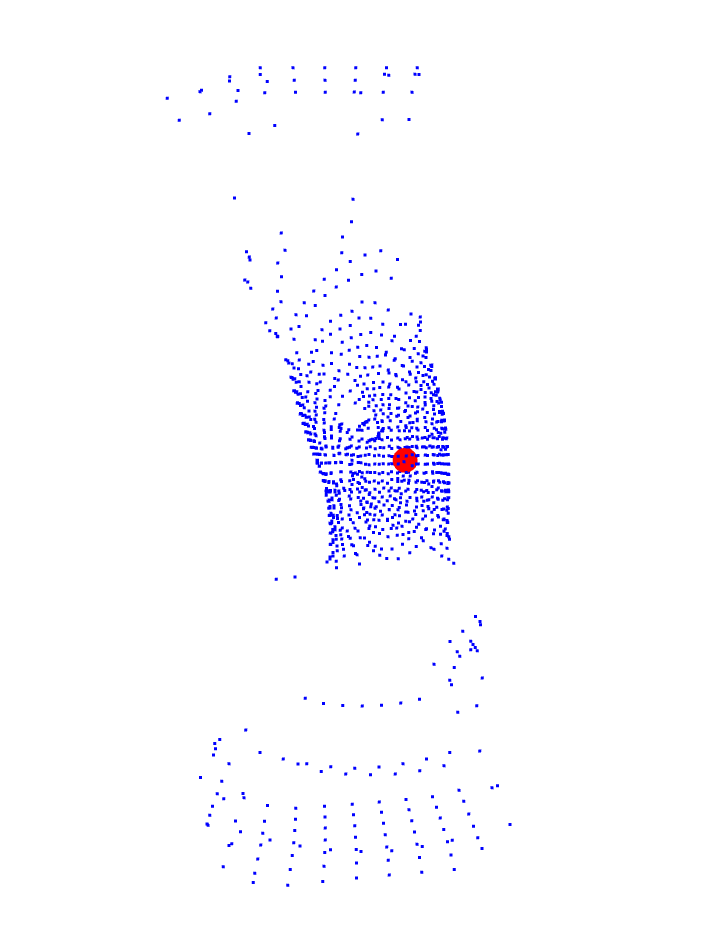}

  \end{subfigure}
  \hfill
  \begin{subfigure}{0.20\columnwidth}
    \includegraphics[width=\linewidth]{figures/screwdriver_fill/1_med_p.png}

  \end{subfigure}
  \hfill
  \begin{subfigure}{0.20\columnwidth}
    \includegraphics[width=\linewidth]{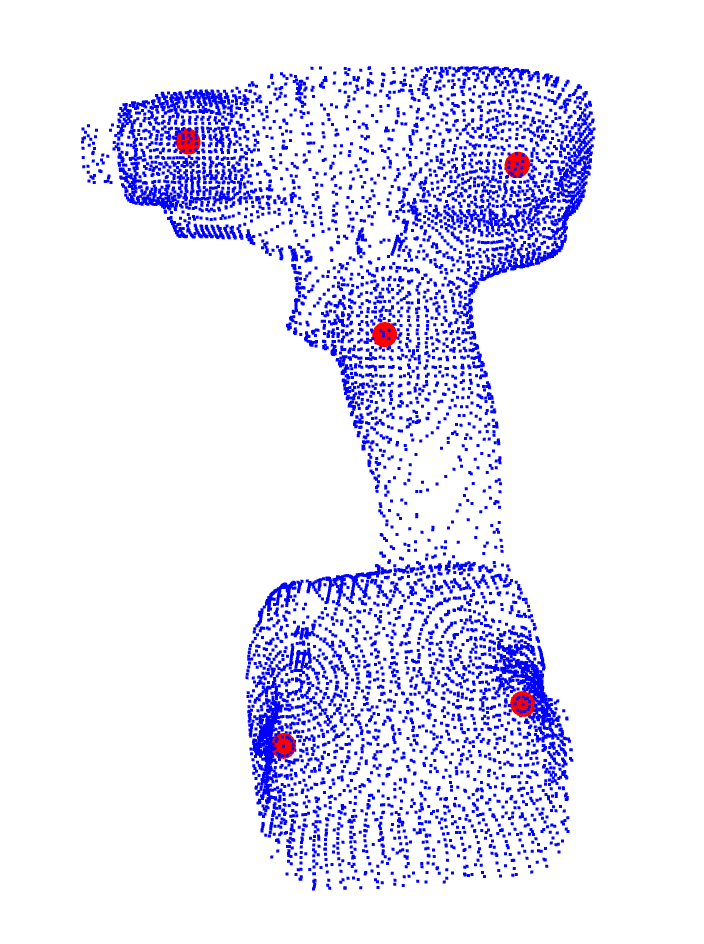}
    
  \end{subfigure}
  \hfill
  \begin{subfigure}{0.20\columnwidth}
    \includegraphics[width=\linewidth]{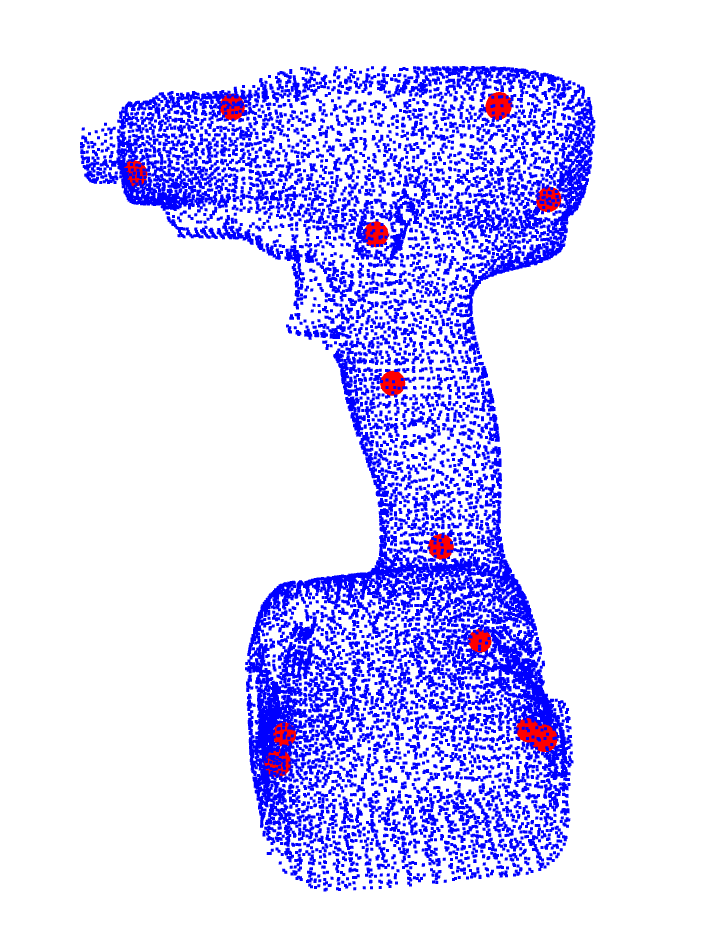}
    
  \end{subfigure}

  \caption{Ray-casting to the objects's surface (blue) from reference points (red) by retaining  the first intersection. Denser coverage achieved with increasing number of reference points (left-to-right).}
  \label{fig:centers}
\end{figure}

\subsection{Surface coverage with mixtures of likelihoods}

It becomes evident that for objects with more complex shapes, a single GP model will not suffice, because there would exist rays emanating from the reference point that intercept the exterior surface of the object in multiple points (e.g. Fig. \ref{fig:centers}). To cope with this issue, we introduce multiple reference points which we distribute preferably in the interior of the object in a way that achieves coverage of the entire surface. Interior reference points ensure that every direction corresponds to a valid training sample. 
A topology-aware placement of such reference points would require distance-based clustering such as k-means 
 ~\cite{lloyd1982least} 
or expectation maximization (EM)~\cite{do2008expectation}, as it will induce a surface partition that will comprise loosely convex patches that are fully reachable from the respective cluster centers via ray-casting. 
For each of these reference points, a new GP prior is introduced. Therefore, the complete object surface is modeled by training multiple predictors of local surface patches with reference points placed at the cluster centers. 

\begin{figure}
  \centering
    \begin{subfigure}{0.24\columnwidth}
         \includegraphics[width=\textwidth]{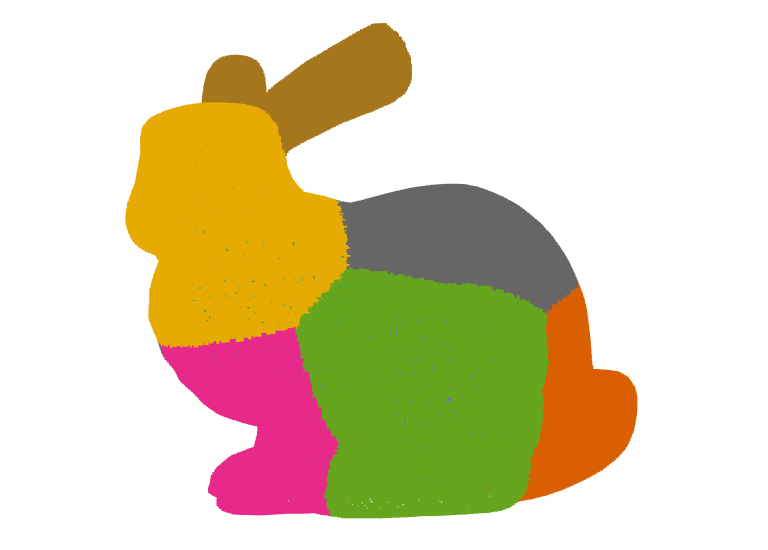}  
         \centering(a)
    \end{subfigure}
    \begin{subfigure}{0.24\columnwidth}
         \includegraphics[width=\textwidth]{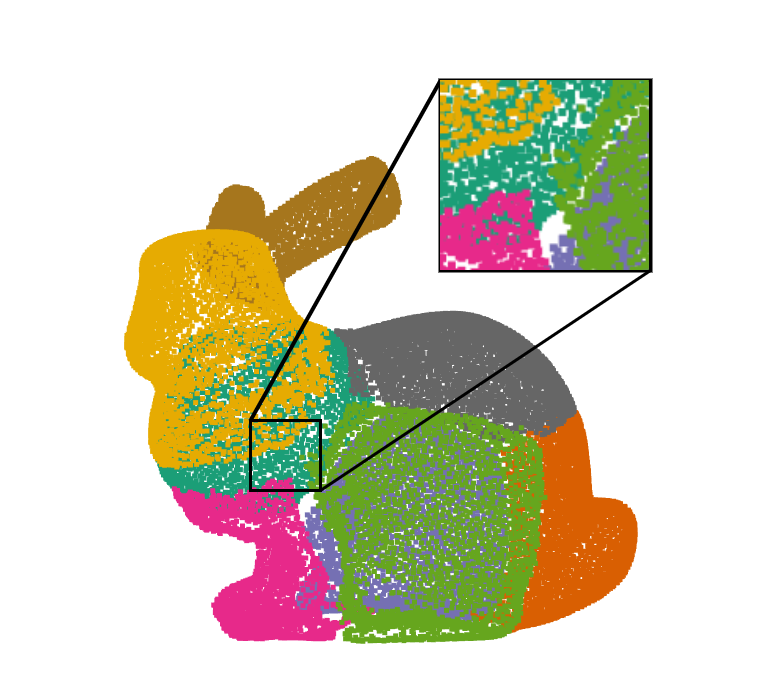} 
         \centering(b)
    \end{subfigure}
    \begin{subfigure}{0.24\columnwidth}
         \includegraphics[width=\textwidth]{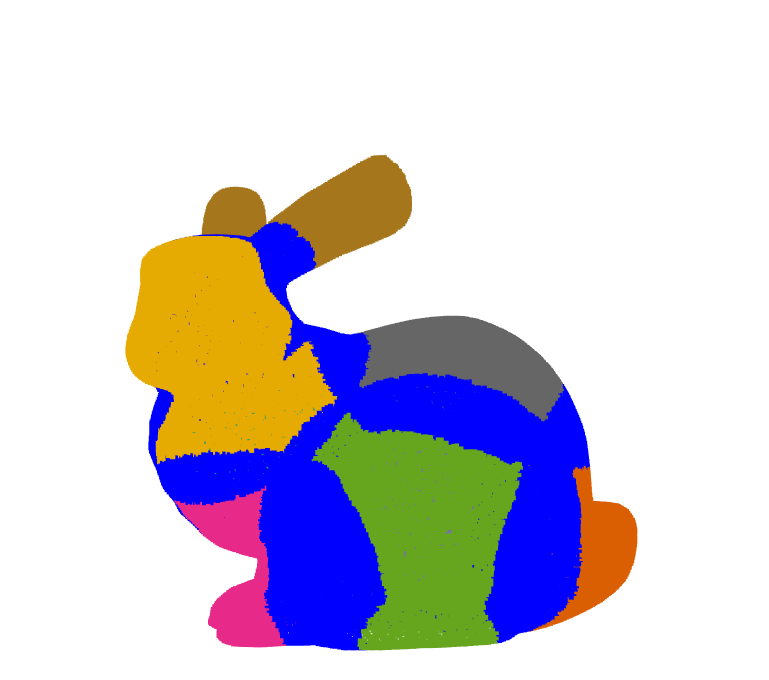}  
         \centering(c)
    \end{subfigure}
    \begin{subfigure}{0.24\columnwidth}
         \includegraphics[width=\textwidth]{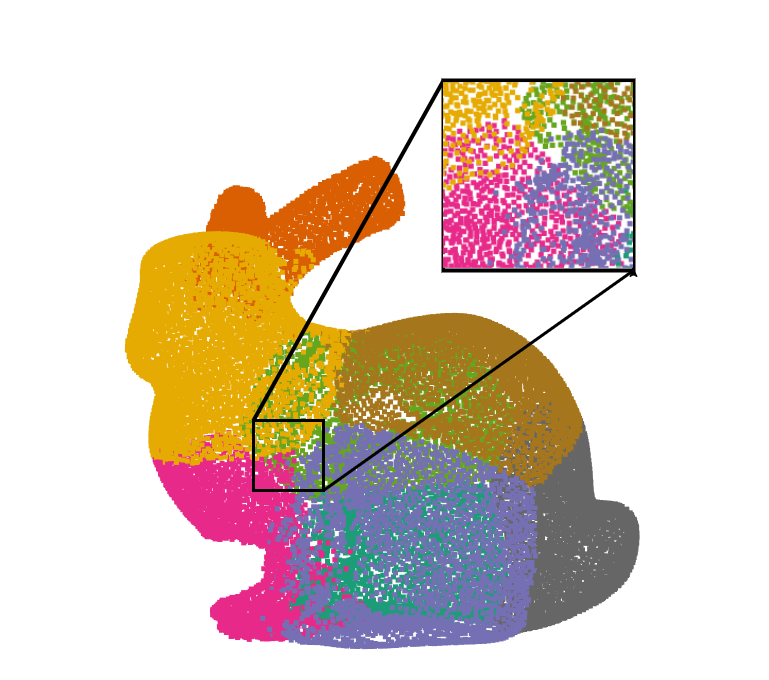} 
         \centering(d)
    \end{subfigure}

  \caption{Assignment of training points to clusters. (a)-(b) Ground truth - reconstructed without overlap. (c)-(d) Ground truth - reconstructed with overlap (blue).}

  \label{fig:bunny_inter_cluster}
\end{figure}

Formally, the idea is to model the likelihood of a 3D point $\pmb{P}$ belonging to the surface of the object $\pmb{o}$ as a mixture of normal distributions derived from the corresponding GP priors with reference points at the cluster centers $\pmb{C}_k$,
\begin{equation}
    p(\pmb{P}\vert \pmb{o})=\sum_{k=1}^K p\left(\pmb{P}\vert\pmb{C}_k; \pmb{o}\right)\pi(\pmb{C}_k; \pmb{o}),
    \label{eq:mixture_model}
\end{equation}
where we choose the likelihood $p\left(\pmb{P}\vert\pmb{C}_k; \pmb{o}\right)$ to be the GP prior based normal distribution of Eq. \eqref{eq:distance_likelihood}, i.e.,
\begin{equation}
    p\left(\pmb{P}\vert\pmb{C}_k; \pmb{o}\right)=q\left(\left\Vert\pmb{P}-\pmb{C}_k\right\Vert \,\big\vert\, \pmb{x}_P, \pmb{C}_k\right),
    \label{eq:mixture_likelihood_from_gp}
\end{equation}
and $\pmb{x}_P$ is the direction parameter vector of $\pmb{P}$ with respect to the reference point $\pmb{C}_k$. Furthermore, the weighting probability $\pi(\pmb{C}_k; \pmb{o})$ is modeled as a softmax ratio based on the distance of $\pmb{P}$ from the reference point $\pmb{C}_k$:
\begin{equation}
    \pi(\pmb{C}_k; \pmb{o})=\frac{\exp\left(-\left(\pmb{P}-\pmb{C}_k\right)^T\pmb{Q}_k\left(\pmb{P}-\pmb{C}_k\right)\right)}{\sum_{l=1}^K \exp\left(-\left(\pmb{P}-\pmb{C}_l\right)^T\pmb{Q}_l\left(\pmb{P}-\pmb{C}_l\right)\right)},
    \label{eq:mixture_softmax_weights}
\end{equation}
where $\pmb{Q}_k$ is a covariance measure of the local space around the k-th cluster center (reference point) $\pmb{C}_k$ \footnote{Provided by the EM algorithm, whereas in the case of k-means we use the identity.}. In practice, this means that we reconstruct a query point $\pmb{P}$ by choosing the nearest center. This however can give rise to borderline errors in the assignment of query points to reference points. To cope with this problem, 
we impose inter-cluster overlap in the training (Fig. \ref{fig:bunny_inter_cluster}).

\begin{table*}[!h]
\centering
\footnotesize 
\setlength{\tabcolsep}{4pt}
\begin{tabular}{l|cccc|cccc|cccc}
\toprule
& \multicolumn{4}{|c|}{\textit{Planes}} & \multicolumn{4}{|c|}{Chairs} & \multicolumn{4}{|c}{Sofas} \\
\cmidrule(lr){2-5} \cmidrule(lr){6-9} \cmidrule(lr){10-13}
& $d_C \downarrow$ & P $\uparrow$ & R $\uparrow$ & F $\uparrow$ & $d_C \downarrow$  & P $\uparrow$ & R $\uparrow$ & F $\uparrow$ & $d_C \downarrow$ & P $\uparrow$ & R $\uparrow$ & F $\uparrow$ \\
\hline
DeepSDF~\cite{park2019deepsdf}    & 0.79  & 90.2  & 85.5  & 87.6  & 0.81  & 84.2  & 88.2  & 86.0  & 0.57  & 89.0  & \textbf{90.0}  & 89.4  \\
DeepSDF (10k)     & 0.80    & \textbf{90.3}  & 84.5  & 87.2  & 0.90  &  84.5 & 88.2  & 86.1  & 0.56  & 65.0  & 60.8  & 62.5  \\
NKSR~\cite{huang2023nksr}     & 0.65  & 78.4  & 81.3  & 79.5  & 0.41  & 84.6  & \textbf{86.3}  & 85.0  & 0.45 & 79.6  & 82.5  & 80.8  \\
\hline
Ours        & \textbf{0.14}  & 87.4  & \textbf{93.6}  & \textbf{90.3}  & \textbf{0.21}  & \textbf{92.3}  & 86.5  & \textbf{89.2}  & \textbf{0.26}  & \textbf{92.4}  & 89.2  & \textbf{90.7}  \\
\bottomrule
\end{tabular}
\caption[Comparison with NKSR and DeepSDF on ShapeNetCore.]{Comparison with NKSR~\cite{huang2023nksr} 
and DeepSDF~\cite{park2019deepsdf} trained with 500K and 10K points per object across shapes of varying complexity for the planes, chairs, and sofas datasets from ShapeNetCore (120 models each). For the NKSR we used the pretrained model on ShapenetCore. We report $d_C$ (Chamfer distance), $P$ (Precision), $R$ (Recall), $F$ (F-score) computed using a sample of 30k points.}
\label{tab:shape_repr_metrics}
\end{table*}

\begin{table*}[htb]
    \centering
    \footnotesize
    \setlength{\tabcolsep}{4pt} 
    \renewcommand{\arraystretch}{1.2} 
    \begin{tabular}{lcccccccccccc}
        \toprule
        & \multicolumn{4}{c}{Planes} & \multicolumn{4}{c}{Chairs} & \multicolumn{4}{c}{Screwdriver} \\ 
        \cmidrule(lr){2-5} \cmidrule(lr){6-9} \cmidrule(lr){10-13} 
        & $d_C$ ↓ & P ↑ & R ↑ & F ↑ 
        & $d_C$ ↓ & P ↑ & R ↑ & F ↑
        & $d_C$ ↓ & P ↑ & R ↑ & F ↑ \\
        \midrule
        \textbf{Polynomial (p=3)} \newline & 1.56 & 35.3 & 58.0 & 43.7 
        & 3.35 & 26.7 & 40.4 & 31.9 
        & 1.69 & 32.6 & 32.8 & 32.7 \\
        \textbf{Periodic} 
        & 0.38 & 70.4 & 86.4 & 77.5 
        & 0.62 & 60.8 & 73.8 & 66.2 
        & 3.47 & 26.6 & 26.8 & 26.7 \\
        \textbf{Linear} 
        & 4.37 & 13.5 & 36.2 & 19.5 
        & 9.09 & 10.9 & 31.8 & 16.0 
        & 9.02 & 10.5 & 14.6 & 12.2 \\
        \textbf{RBF} 
        & 0.31 & 73.0 & 88.9 & 80.1 
        & 0.47 & 60.7 & 69.2 & 66.4 
        & 0.20 & 86.0 & 88.4 & 87.2 \\
        \textbf{RQ} 
        & \textbf{0.19} & \textbf{82.6} & \textbf{93.8} & \textbf{87.9} 
        & 0.51 & \textbf{68.4} & \textbf{82.9} & \textbf{74.5} 
        & \textbf{0.12} & \textbf{95.1} & \textbf{96.5} & \textbf{95.8} \\
        \textbf{Matern} 
        & 0.26 & 76.3 & 91.8 & 83.3 
        & \textbf{0.43} & 64.7 & 79.2 & 70.8 
        & 0.15 & 92.0 & 93.6 & 92.8 \\
        \bottomrule
    \end{tabular}
    \caption{Ablation study of different kernels and their impact on reconstruction accuracy across various shapes. Chamfer distance is multiplied by $10^3$.}
    \label{tab:kernel_choice}
\end{table*}

\subsection{Experimental setup}
\subsubsection{Datasets} We conduct our experiments on two datasets: ShapenetCore~\cite{shapenet2015} and IndustryShapes \cite{sapoutzoglou2026industryshapes}. ShapeNetCore is a densely annotated subset of the full ShapeNet corpus, containing 51,300 unique 3D models that span across 55 common object categories. We limit our experiments to 3 representative object categories, namely \textit{Planes}, \textit{Chairs} and \textit{Sofas} and use 120 models per category. IndustryShapes is a domain-specific dataset containing a limited set of tools and components that are used in an industrial setting of car-door assembly. For our experiments, we use the \textit{Screwdrivers} object category. 

\subsubsection{Evaluation metrics} To demonstrate the accuracy of our GP mixture-based modeling for surface reconstruction we report the following metrics: 

\PAR{Chamfer distance ($d_{C}$)}. Measures the similarity between the ground-truth and estimated point sets $\mathbf{P_{gt}, P_{est}}$, and is defined as:

\begin{equation}
\begin{split}
     \mathbf{d_C}(P_{gt}, P_{est}) = \frac{1}{|P_{gt}|} \sum_{x \in P_{gt}} \min_{y \in P_{est}} \|x - y\|^2 + \\ \frac{1}{|P_{est}|} \sum_{y \in P_{est}} \min_{x \in P_{gt}} \|x - y\|^2.
\end{split}
\end{equation}

\PAR{Precision (\textit{P}).} Measures the accuracy of the predicted points by quantifying how closely they match the ground truth.
We compute the minimum distance between each predicted point $p_{est}$ and the ground truth set $P_{gt}$:
$$d_{p_{est} \rightarrow {P_{gt}}} = \min_{p_{gt} \in P_{gt}} \|p_{est} - p_{gt}\|.$$
By defining a distance threshold 
$\tau$, we can determine the percentage of predicted points whose distance to the ground truth falls within this threshold: 
\[
\mathbf{P}\left(\tau\right) = \frac{|\{p_{est} \in P_{est} :d_{p_{est} \rightarrow {P_{gt}}} < \tau\}| }{|P_{est}|}.
\]
In our experiments, we used a threshold of $\tau$ = 0.01 as all models are normalized to the unit sphere.

\PAR{Recall (\textit{P}).} Measures the completeness of the reconstruction, quantifying how well the ground truth points are represented in the predicted point cloud. Similar to Precision, we compute the minimum distance between each ground truth point $p_{gt}$ and the predicted set $P_{est}$:
$$d_{p_{gt} \rightarrow {P_{est}}} = \min_{p_{est} \in P_{est}} \|p_{gt} - p_{est}\|.$$
Using the same distance threshold $\tau$, Recall is the percentage of ground truth points that are within this threshold from the predicted points:
\[
\mathbf{R}\left(\tau\right) = \frac{|\{p_{gt} \in P_{gt} :d_{p_{gt} \rightarrow {P_{est}}} < \tau\}| }{|P_{gt}| }.
\]
We used the same distance threshold of $\tau$ = 0.01 for recall as well.

\PAR{F-score (\textit{F}).} The F-score combines Precision and Recall into a single metric, providing a more robust evaluation than either precision or recall alone. The F-score is defined as:

$$\mathbf{F} = \frac{2 \mathbf{P}(\tau) \mathbf{R}(\tau)}{\mathbf{P(\tau)} + \mathbf{R(\tau)}}.$$

\subsubsection{Data preparation} We prepare our input data following the methodology of DeepSDF~\cite{Park_2019_CVPR} to allow for a fair comparison. In particular, we first normalize the input model to the unit sphere. Subsequently, we sample a dense point cloud from the model's mesh surface by casting rays from a set of virtual cameras placed on a Fibonacci grid~\cite{swinbank2006fibonacci} onto the surface of the sphere. Our training and testing input is formed by sub-sampling two completely distinct sparse point clouds containing 10,000 and 30,000 points respectively.

\subsubsection{Optimization} Our method uses GPyTorch~\cite{gardner2018gpytorch} for training and evaluating the GP models. We use the Adam optimizer~\cite{kinga2015method} with an initial learning rate of 0.1 and employ a learning rate scheduler to dynamically reduce the learning rate by a factor of 0.1 after 10 iterations without improvement in the loss function. All experiments were conducted using a single RTX 4090 GPU with 24GB of VRAM. For an input sparse point cloud of 10,000 points, the GP models typically take about 30 seconds to train on the object, with slight variations depending on the number of reference points. Introducing overlapping regions increases the overall number of training samples.

 \begin{figure*}[!h]
    \centering
    \includegraphics[width=0.9\textwidth]{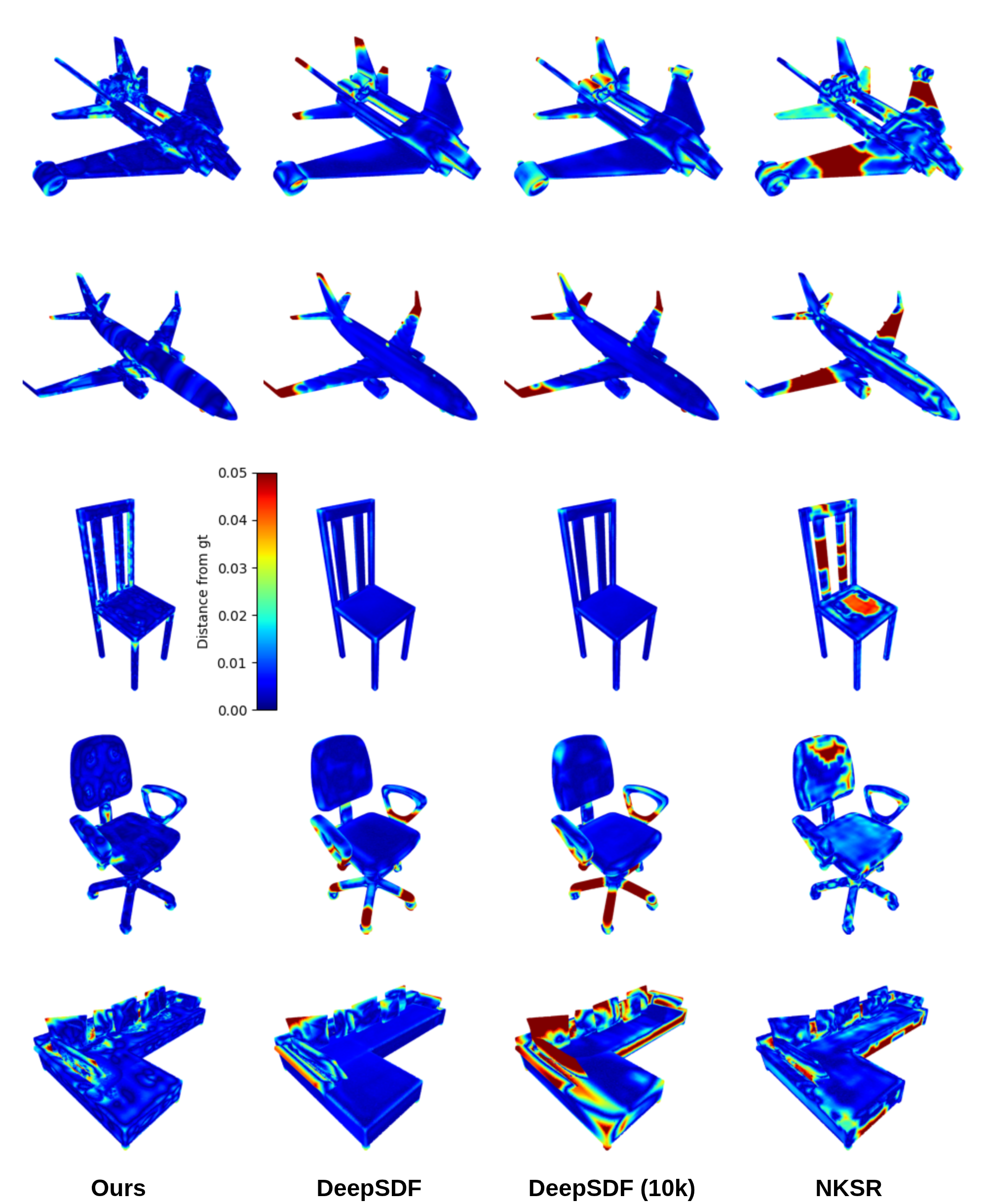}
    \caption[Qualitative comparison of shape representation on the ShapeNetCore dataset]{Qualitative comparison of shape representation on the ShapeNetCore dataset. The heatmap quantifies the distance from the ground truth point-cloud to the reconstructed; superimposed into the ground truth point cloud. We demonstrate two examples from the airplanes, chairs and sofas categories with each method. From left to right ours, DeepSDF~\cite{park2019deepsdf}, DeepSDF (10k), NKSR~\cite{huang2023nksr}}
    \label{fig:heatmap_results}
\end{figure*}

\begin{figure*}[htb]
    \centering
    \includegraphics[width=0.8\linewidth]{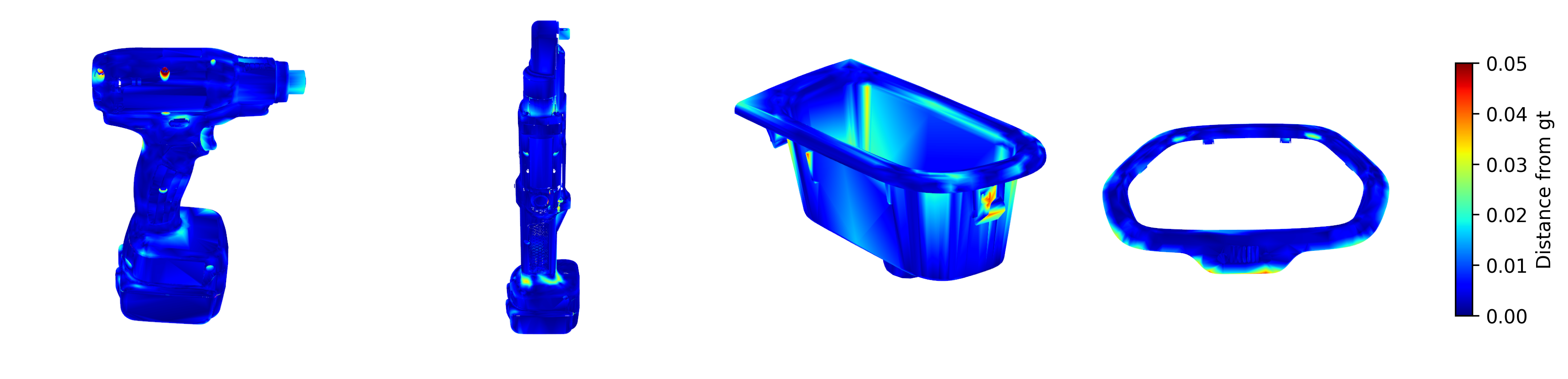}
    \caption{Qualitative results of our method for the models of the IndustryShape dataset.The heatmap quantifies the distance from the
ground truth point-cloud to the reconstructed; superimposed into the ground truth point cloud.}
    \label{fig:industry_shapes_heatmap}
\end{figure*}

 \subsection{Main results}
 \label{sec:eval_analysis}

Table \ref{tab:shape_repr_metrics} presents a quantitative comparison between our proposed shape representation and two state-of-the-art competitors: DeepSDF~\cite{park2019deepsdf} and NKSR~\cite{huang2023nksr}. Our method achieves the lowest Chamfer distance and over 90\% F-score across the board. Similarly, competitive Precision and Recall are attained for all object categories, with a remarkable 93.6\% Recall for the \textit{Planes} class. As the qualitative comparison of Fig. \ref{fig:heatmap_results} illustrates, our representation is able to capture the majority of the shape details in a robust manner, while competitors either over-smooth important shape features or miss them completely. It should be noted that we follow an intrinsically different approach. While DeepSDF and NKSR focus on generalization across different objects, the former by learning the latent space of shapes and the latter by combining the conditioning of kernel parameters on the data with kernel ridge regression, our method is more adept in representing details of specific objects. Fig. \ref{fig:industry_shapes_heatmap} illustrates the reconstruction results of our representation method applied to models from the IndustryShapes dataset. As demonstrated, our method successfully models complex, thin geometries and preserves challenging topological features, such as holes, to a significant extent.

\subsection{Ablation studies}
\label{sec:res_abl}

\subsubsection{Reference points.} 
\begin{figure}[H]
  \centering 
  \begin{subfigure}[t]{0.28\columnwidth}
      \includegraphics[width=\linewidth]{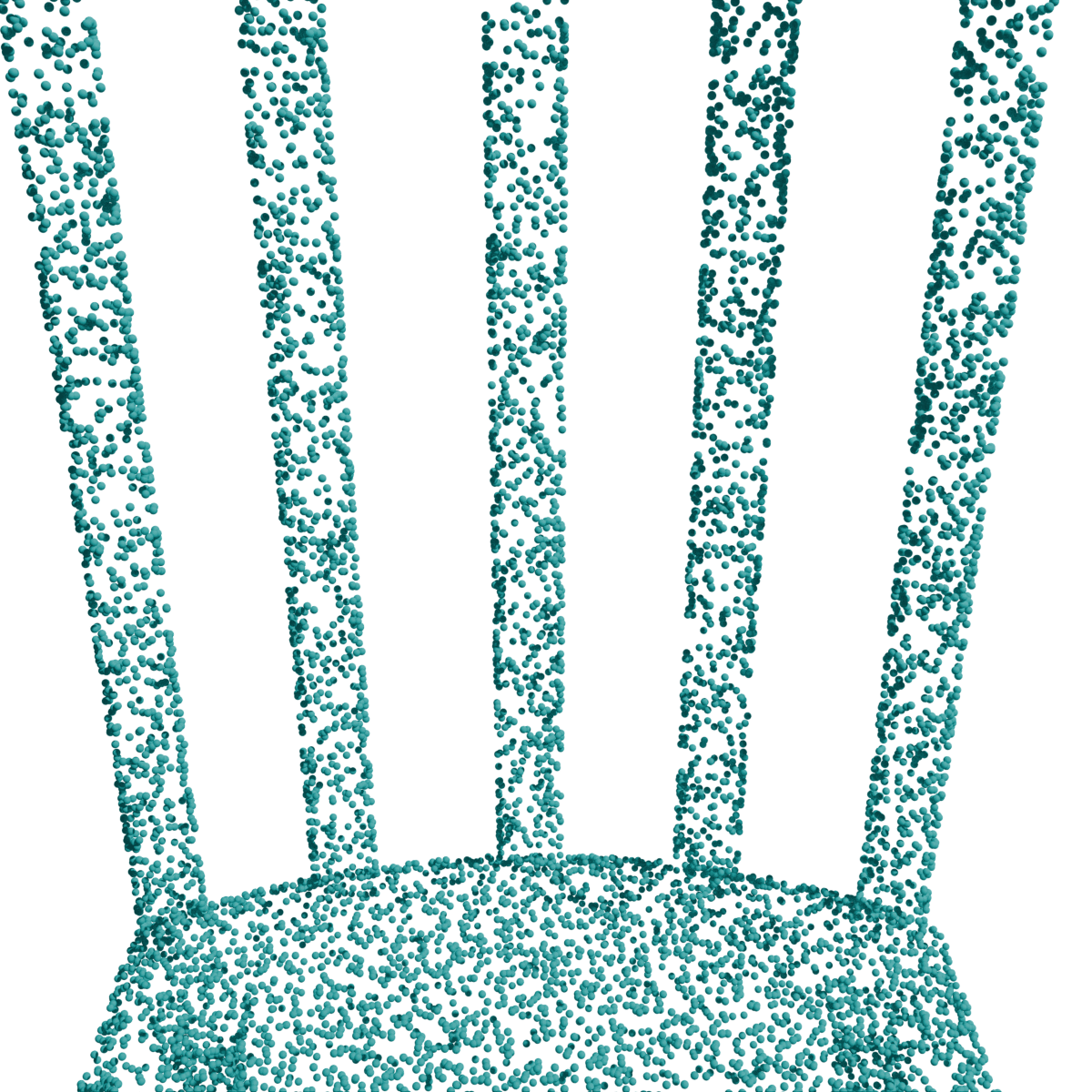}
    \caption{}
  \end{subfigure}
  \begin{subfigure}[t]{0.28\columnwidth}
    \includegraphics[width=\linewidth]{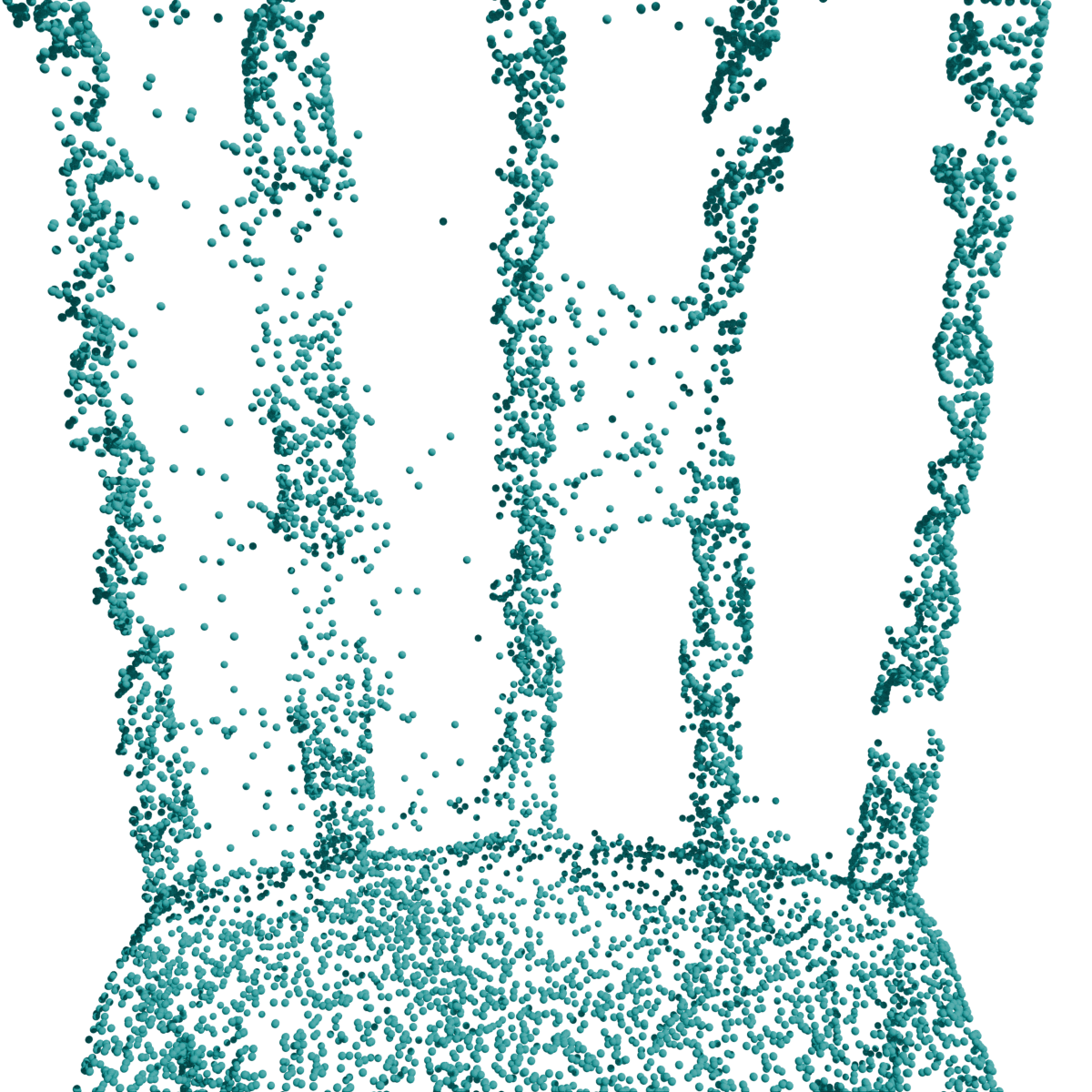}
    \caption{}
    \label{subfig:chair_e}
  \end{subfigure}
  \begin{subfigure}[t]{0.28\columnwidth}
    \includegraphics[width=\linewidth]{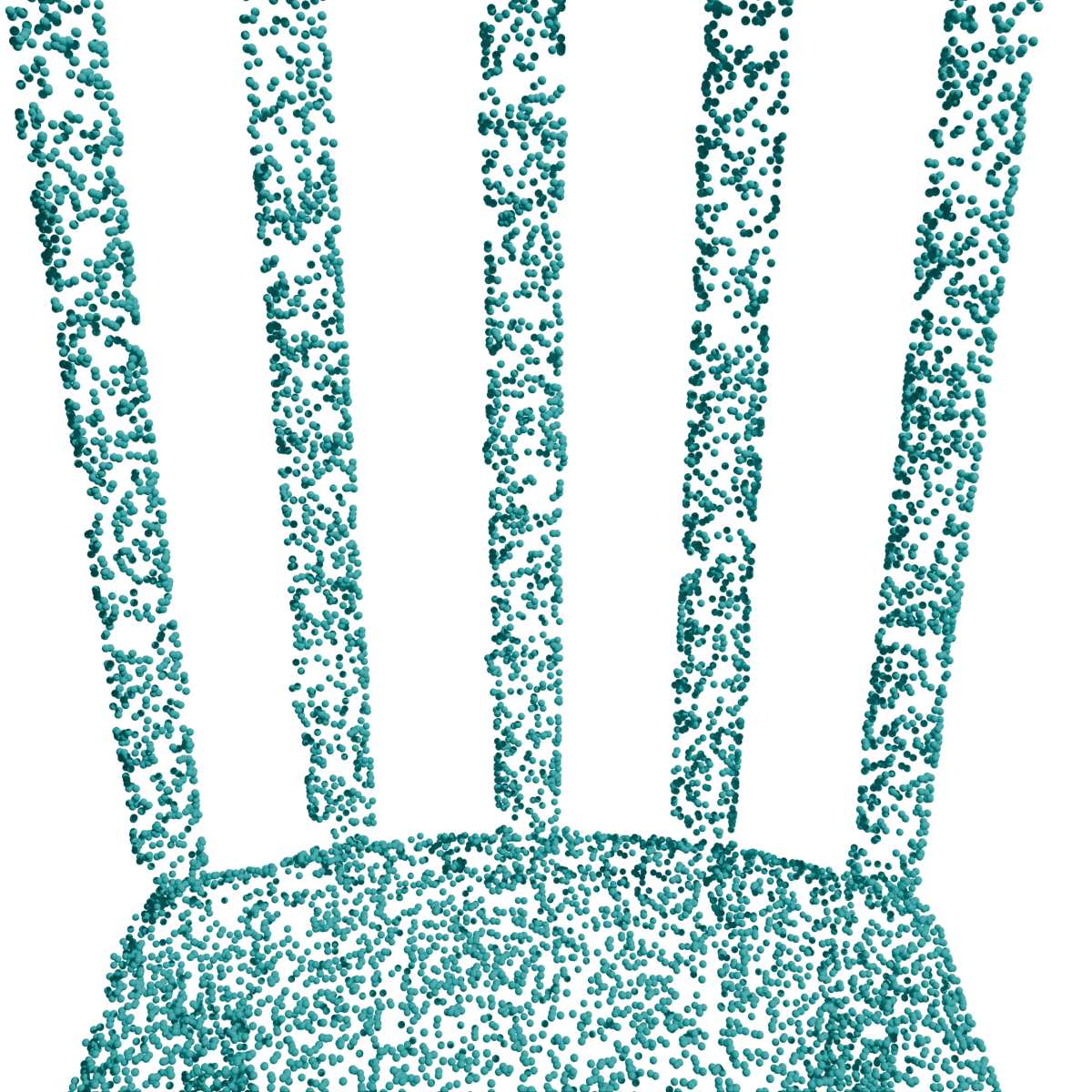}
    \caption{}
    \label{subfig:manual_chair_centers}
  \end{subfigure}

  \caption{(a) Ground truth, (b) Results obtained from automatically estimating reference point locations (16) with k-means, (c) Results based on manually defined reference points (32).}
  \label{fig:chair_manual_automatic}
\end{figure}

\begin{figure}[htb]
\centering
  \includegraphics[width=0.9\columnwidth]{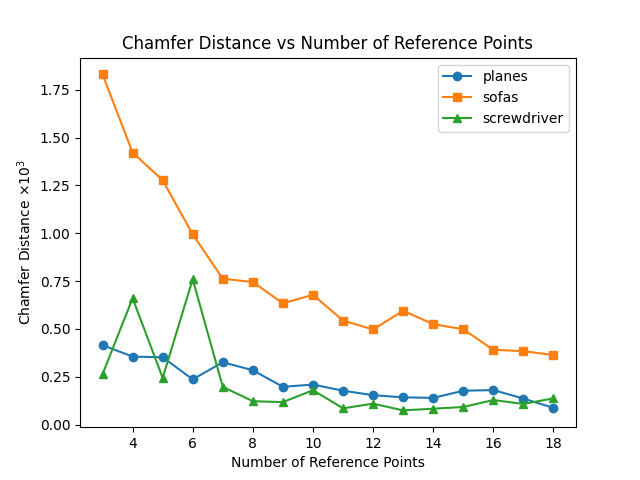}
  \caption{Ablation study for various number of reference points vs shape accuracy based on CD for the planes and sofas object classes of the \emph{ShapeNetCore} dataset and the screwdriver from the IndustryShapes dataset.}
  \label{fig:ablation}
\end{figure}

We first study the impact that the multitude and the positioning of the reference points might have on the template's fidelity. To isolate the effect of reference point quantity, Fig.~\ref{fig:ablation} illustrates the Chamfer distance trends for the planes, sofas, and screwdriver classes. For these categories—where automatic distance-based clustering (k-means) was used—increasing the number of reference points generally improves representation quality until reaching a plateau. However, for highly complex topologies with multiple tube-like structures, such as the chairs category, k-means initialization lacks the semantic information needed to efficiently place reference points at strategic locations. Therefore, rather than evaluating the chairs category based on quantity alone, we utilize it to demonstrate the critical importance of ideal reference point placement. This specific limitation highlights the necessity of employing more intelligent algorithms, such as skeletonization, to compute reference centers for complex shapes. As shown in Fig.~\ref{fig:chair_manual_automatic}, automatic distance-based placement is inferior to manual placement for such complex objects, though it remains adequate for capturing the underlying topology (see also Sec.~\ref{sec:res_limits}).

\vspace{-0.6cm}
\subsubsection{Kernel choice.} To identify a suitable kernel that performs robustly across objects of varying classes and complexities, we conducted structured experiments testing a range of kernel types~\cite{Rasmussen2004}, ranging from a third-degree polynomial to periodic, linear, RBF, Matérn, and Rational Quadratic (RQ). To ensure a fair comparison, all models were evaluated using eight fixed reference points per object. For this ablation, we specifically evaluated five models each from the Planes and Chairs categories alongside two models from the IndustryShapes Screwdrivers class. These categories were deliberately selected over others (such as Sofas) because they present highly contrasting and challenging geometric features: Planes contain thin, protruding structures, while Chairs possess complex topologies with high-frequency details and tube-like structures. The mean metrics for each category (Chamfer distance, Precision, Recall, and F-score) are detailed in Table \ref{tab:kernel_choice}. The results indicate that while the RBF and Matérn kernels perform well, the RQ kernel exhibits superior overall performance and adaptability across the tested object categories.


\subsection{Limitations}
\label{sec:res_limits}
Aside the number of points that impact the template fidelity (cf.~Sec.~\ref{sec:res_abl}), the locations of the reference points may also influence the quality of the representation. Overall, perturbing the number of reference points and their locations could lead to improvements, particularly when dealing with complex objects. For instance, in the case of the chair depicted in Fig.~\ref{fig:chair_manual_automatic}, comparing results obtained from clustering-based estimation of reference points (cf. Sec.~ \ref{ssect:lightweighht_shape_templates}) with those manually defined, a more accurate shape representation is derived for the latter.

\section{Conclusion \& Future work}

This work presents a functional, probabilistic shape representation grounded in Gaussian-Process mixtures over directional distance fields. The method provides a lightweight, interpretable, and high-accuracy alternative to neural implicit functions, especially effective under sparse sampling. In future work, we aim at
endowing the shape representation algorithm with theoretical guarantees for full coverage of the object surface irrespective of shape complexity via learned mixtures models and fully automating the choice of reference points.

\section*{Acknowledgments}
This work was partially funded by the European Union’s Horizon Europe programme SOPRANO (GA No 101120990) and PANDORA (GA No 101135775).

{
	\begin{spacing}{1.17}
		\normalsize
		\bibliography{ISPRSguidelines_authors} 
	\end{spacing}
}

\end{document}